\documentclass[final,5p,times,twocolumn]{elsarticle}

\usepackage{graphicx}
\usepackage{lscape}
\usepackage{verbatim}
\usepackage{algorithmic}
\usepackage{algorithm2e}
 
\usepackage{amsmath}
\usepackage{amssymb} 
\usepackage{subfigure}

\usepackage{tabularx,ragged2e,booktabs,caption,array,multirow,multicol}
\usepackage{csquotes}
\usepackage{mathtools} 

\usepackage{lineno}
\usepackage{amsthm}  
\usepackage{listings}
\usepackage{amssymb}
\usepackage{latexsym}
\usepackage{epsfig}
\usepackage{float}
\usepackage{xspace}
 \usepackage{float}

\usepackage[colorlinks=true,linkcolor=blue,citecolor=blue,pdfmenubar=true]{hyperref}

   \graphicspath{{../pdf/}{../jpeg/}}
  \DeclareGraphicsExtensions{.pdf,.jpeg,.png}

\journal{ }

\begin{document}

\begin{frontmatter}

\title{   An affective computational model for   machine 
consciousness  }


 \author{Rohitash Chandra }

\address{   Artificial Intelligence and Cybernetics Research Group, Software 
Foundation, Nausori, Fiji\\}

\begin{abstract}

In the past, several models of consciousness have become popular and  have led 
to the development of models for machine consciousness with 
varying degrees of success and challenges for simulation and implementations. 
Moreover, affective computing attributes that involve emotions, behavior and 
personality have not been the focus of models of consciousness as they lacked 
motivation for deployment in software applications  and robots.    The 
affective 
 attributes are important factors for the future of machine consciousness with 
the rise of  technologies  that can assist humans. Personality and affection 
hence can give an additional flavor for the computational model of  
consciousness in 
humanoid robotics. Recent advances in areas of machine learning with a focus on 
deep learning can further help in developing aspects of machine consciousness 
in 
 areas that can better replicate human sensory perceptions such as speech 
recognition and  vision. With such advancements, one encounters further 
challenges in developing models  that can synchronize different aspects of 
affective computing.  In this 
paper, we review some existing   models of 
consciousnesses and present an
affective computational model   that would enable the human touch and feel for 
robotic systems.


\end{abstract}


 \end{frontmatter}


\section{Introduction}


The   definition of  consciousness  has been a  major 
challenge  for  simulating or modelling 
human 
consciousness \cite{Chalmers1995,Searle1997}.   Howsoever,  a broad 
definition of consciousness is the state or quality of awareness which features 
sentience,  subjectivity, the ability to experience through sensory perceptions, 
the state of wakefulness,   the sense of ego, and  the  control  of the mind 
with 
awareness of thought processes 
\cite{baars1997theatre,Haarmann1997,revonsuo2006inner,Wallace2005,
Haladjian2016210}.




%

The challenges in the definition and models of consciousness affects the 
implementation or simulation study  of consciousness. In the past, simulation 
study has been presented for certain models of 
consciousnesses, such as the  model of information flow from global workspace 
theory\cite{baars1997theatre}. Shanahan presented a study where cognitive 
functions such as anticipation and planning were  realised through internal 
simulation of interaction with the environment.  An implementation    
based on weightless neurons was used to control a simulated robot 
\cite{shanahan2006simula}. Further 
attempts were made to model specific forms of intelligence through 
brute-force search heuristics to  reproduce features of human perception 
and cognition, including emotions \cite{Haladjian2016210}. 

Moreover,   small 
scale 
implementations   can consider models based from consciousness in 
animals that are needed for their survival.  Although 
intelligence demonstrated in  solving the tasks vary  
\cite{griffin2013animal,panksepp2005affective}, limiting definitions of 
consciousness to humans is 
speculative as all living beings tend to have certain attributes that overlap 
with human consciousness.   Some of the 
undomesticated animals  such as  rodents  have a history of survival in 
challenging and wide range 
of climate and environments \cite{callaghan2012effect}. There are 
some studies that show that  animals such as rats  seem to express some 
aspects of consciousnesses, that is not merely for survival. They
 feature   social attributes 
such as empathy which is similar to humans  \cite{Bartal1427,Langford1967}. 
High level of 
curiosity and  creativity  are  major attributes of 
consciousness 
which  could be a factor that distinguishes humans from rest of the animals 
\cite{Preiss2016,Hunter201648,Lin201636}. While intelligence is also an 
underlying aspect of consciousness, it has been shown that intelligence is a 
necessary, however, not sufficient condition for creativity 
\cite{Karwowski2016105}. Howsoever,  apart from humans, other animals  also 
show certain levels of creativity \cite{kaufman2011towards}. There has been 
attempts to enhance existing models in unconventional ways 
through means  to incorporate  non-materialistic aspects of 
consciousnesses through studies of near-death experiences 
\cite{Schwartz2015252}. Furthermore, ideas from psychology and quantum
mechanics have also been integrated in a study  that
challenge the materialistic view of consciousness \cite{Brabant2016347}.

 

 In an attempt to empirically study  consciousness, Tononi proposed 
the information integrated theory of consciousness  to quantify the amount 
of integrated information an entity possesses  which determines its level of  
consciousness \cite{tononi2004information}. The theory  depends exclusively on 
the ability of a system to integrate information, regardless of having a strong 
sense of self, language, emotion, body, or   an environment. Furthermore, it 
attempts to  explain why consciousness requires neither sensory input nor 
behavioural output in cases such as  during  the sleeping state. Further work 
was done with application of integrated information to discrete networks as a 
function of their dynamics and causal architecture 
\cite{balduzzi2008integrated}. Information integrated theory 3.0 further 
refined the properties of consciousness with phenomenological axioms and  
postulates to lay out  a system of mechanisms to satisfy those axioms and thus 
generate consciousness   \cite{oizumi2014phenomenology}. It was suggested that 
systems with a purely feed-forward architecture cannot generate consciousness, 
whereas
 feed-back  or  recursion  of some nature could be an essential 
ingredient of consciousness. This was based on a previous study, where it was 
established 
 that the presence or absence of feed-back could be directly equated with
the presence or absence of consciousness \cite{lamme2003visual}. 

David Chalmers  
highlighted the explanatory gap in defining consciousness and indicated that 
the hard problem of consciousness
 emerge from    attempts that try to  explain it in purely physical 
terms \cite{Chalmers1995}. Integrated information theory is based on 
phenomenological axioms which  \textit{begins with consciousness}  and  
indicates 
that complex systems with   some feedback 
states could have varying levels of consciousness 
\cite{oizumi2014phenomenology}. Howsoever, this does not fully support the 
motivations for consciousness experience as defined by Chalmers that look at 
 conscious experience or \textit{qualia} from  first and third person 
perspectives and the relationship between them \cite{chalmers2013can}.



The field of affective computing focuses on the 
development of systems that can simulate, recognize,   and process human  
affects which   essentially is   the experience of feeling or emotion 
\cite{picard2010affective,picard2003affective,picard1997affective}. Affective 
computing could provide  better  communication  between humans and 
artificial systems that can lead to elements of   trust and
connectivity with artificial systems \cite{tao2005affective}. The motivation to 
have affective models in artificial consciousness would be 
towards the future of mobile technologies and robotic systems that guide in 
everyday human activities. For instance, a robotic system which is part of the 
 household kitchen could  further feature
communication that builds  and connectivity from features of affective 
computing  \cite{yamazaki2010recognition}. In the near future, there 
will also be a growing 
demand for sex robots, therapeutic and  nursing robots  which would  
need affective computing features  \cite{Bendel2015,sharkey2012granny}. 
Moreover, the emergence 
of smart toys and robotic pets could be helpful in raising  children and 
also assist  the elderly \cite{sharkey2012granny}.  Although 
mobile application-based support  and learning systems have been 
successfully deployed, they are often criticized for having less physical 
interactions \cite{mira2014spanish}.  In such areas, affects in robots   could 
lead to further help such as stress management and  counselling.

Personality is an integral part of consciousness \cite{barrick1991big}. However, 
in 
the past, the proposed models of consciousness have not tackled the feature of  
personality  \cite{costa1994personality}. In the past, a study 
presented the influence of different types of personality on work performance 
for  selection, training and development, and performance appraisal  of workers 
\cite{barrick1991big}. Nazir et. 
al further presented culture-personality based affective model that included  
the 
 five dimensions of personality  \cite{nazir2009culture}. Carver  and 
Scheier used   control theory as a conceptual framework for personality which 
provides an understanding of social, clinical and health psychology  
\cite{carver1982control}.   Although these studies have been very popular in 
areas of psychology, there has not been much work done to incorporate 
understanding of personality in models of machine 
consciousness. 
 

We note that element of hunger and pain are 
some of the leading biological attributes for survival which contributes to 
human personality and affects. Starzyk et. al 
presented motivated learning for the development of autonomous systems  
based on competition between dynamically-changing pain signals which provided  
an interplay of externally driven and internally
generated control signals \cite{starzyk2012motivated}. The use of abstract 
notion of pain as a motivational behaviour for a goal such as food can lead to 
features in affective model for machine consciousnesses.  Although several 
prominent models of machine  consciousness have been present, 
their limitations exist in terms of addressing  the  features of  human affects 
that could lead future 
implementations in robot systems and other related  emerging technologies. In 
such systems with human 
affects, there would be a wider impact in terms of social acceptance, trust and 
reliability. However, the limitations that exist in humans could also pose a 
threat. We limit our 
focus on the development of affects that could lead to personality in 
artificial 
consciousness without much emphasis for  implications or social 
acceptable of such systems.

   In this 
paper, we review some existing   models of 
consciousnesses and present an
affective computational model of machine  consciousness with the motivation to 
incorporate human affects and  personality. We promote a 
discussion of using  
 emerging technologies  and advances in machine  learning  for developing the 
affective computational model.

The rest of the paper is organised as follows. Section II provides a background 
on consciousness and existing models. Section III presents the 
proposed model and Section IV provides a discussion with further research 
directions while Section V concludes the paper. 

 
\section{ Background and  Related Work} 
 
 

 \subsection{Studies  of consciousnesses}


Although certain foundations in the definition of consciousness have 
emerged \cite{baars1997theatre,Haarmann1997,Wallace2005}, there has been the 
need for a definition that can 
fulfill the needs from the perspective of various fields that include 
neuroscience, psychology and 
philosophy. Historically, the study of consciousnesses has been 
the subject of various groups  and phases in ancient and modern history that 
include those both from Eastern 
\cite{billington2002,dasgupta1922history} and Western philosophical traditions 
\cite{russell2013history}. 

There are some difficulties in defining consciousness that led to identifying 
areas known as the easy and the  hard 
problems of  consciousness \cite{harnad2000mind,gray2004hardproblems} from 
perspectives of neurobiology and neurophilosophy.  
Chalmers introduced the hard problem which highlights the explanatory gap of  
defining the conscious experience, though which  
sensations acquire characteristics, such as colors and taste 
\cite{Chalmers1995}.   The rest of the problems are the 'easy 
problems' that generally refer to the functions
such as accessibility and reportability, howsoever, they are also unsolved 
problems in cognitive science \cite{Chalmers1995,chalmers2013can}. The  easy 
problems of consciousness constitute of  the ability to discriminate, integrate 
information, report mental states, and focus attention.  These could be  
deduced and modeled through advances in  artificial intelligence 
\cite{mcdermott2007artificial}. Chalmers also proposed a pathway 
towards the science of understanding consciousness 
experience through the integration of  third-person data about 
behavior and brain processes with first-person data about conscious experience 
\cite{chalmers2013can}.  Moreover, the easy problems in consciousness could be 
tackled by constructs in weak artificial 
intelligence (AI) \cite{searle1980minds}. Note that strong AI refers to the 
notion that machines can think similar to humans and possess some level 
of consciousness and  sentience,  while weak AI refers  to machines that can be 
made to act as if they are intelligent  \cite{searle1980minds}.


 There have been also been concerns about the 
 ability of neuroscience to explain properties of consciousness 
 \cite{brain1956cons,Chalmers1995,crick1995neuroscience}. Chalmers argued that 
neuroscience is good in explaining easy problems of consciousness and
faces 
 major challenges in the hard problems \cite{Chalmers1995}. The 
mind-body problem is   one of the historical challenges  about the 
nature of 
consciousness \cite{mcginn1989can}. In this problem, there is  dilemma   about 
the relationship of the mind with  the brain  since mental states and processes 
such as thinking are non-physical while the human body is a physical entity 
\cite{bunge2014mind,kandel2007search}.  Uncertainties in definitions of 
consciousness \cite{Searle1997} have been also 
promoting views of consciousness that have been more metaphysical  and 
spiritual \cite{tola2004being,shusterman2008body}.  There has also been 
evidence of consciousness related abnormalities in physical systems suggesting  
that consciousness  can  alter the outcome of certain physical  processes such 
as random-number generators \cite{radin1989evidence}. Although 
these topics are interesting, 
certain restrictions need to be placed in the development of machine
consciousness that can lead to the development of robotics and other related 
intelligent systems that can assist in human decision making and also carry out 
everyday tasks. We, therefore, limit out definition of consciousness merely to 
that which can help in the formulation of problem-solving techniques,  which 
may 
restrict to models relation to information theory of consciousness 
\cite{tononi2012integrated} that could lead to software systems or models that 
to replicate consciousness to a certain degree.

 One of the issues of the mind-body problem has been in the explanation of the 
links that govern the physical (brain) with the non-physical (mind). This can 
be seen as analogous to the relationship between hardware and  software 
systems.  Wang presented a study with  a comprehensive set of informatics and 
semantic properties and laws of software as well as their mathematical models 
\cite{wang2006informatics}. In order to provide a rigorous mathematical 
treatment of both the abstract
and concrete semantics of software, a new type of formal semantics known as the 
deductive semantics was developed. Later,  a 
theoretical framework of  cognitive informatics  that was shown to be a 
trans-disciplinary inquiry of the internal information processing mechanisms 
and processes of the brain and natural intelligence \cite{wang2008theoretical}. 
Furthermore, Wang et. al presented an architecture, theoretical foundations,
and engineering paradigms of contemporary cybernetics with a link to 
computational intelligence has been introduced in the
cybernetic context and the compatibility between natural and cybernetic 
intelligence was analyzed\cite{wang2009contemporary}.  Moreover, Wang presented 
a formal model and
a cognitive process of consciousness in order to explain how abstract 
consciousness is generated. The hierarchical levels of consciousness were
explored from the facets of neurology,
physiology, and computational intelligence. A rigorous mathematical 
model of consciousness was  created and the cognitive process of 
consciousness is formally described using
denotational mathematics \cite{wang2012cognitive}
 
 \subsection{Existing models for machine consciousness}
 
 Over the last few decades, various attempts have been made to use studies of 
consciousnesses for models or development of machine consciousness. While there 
are various models with certain strengths and limitations, in general, there 
lacks simulation study for these models. Gamez initially presented a review of 
the progress in machine consciousness where the literature was divided 
into four groups that considered of  the external behavior,   cognitive 
characteristics,   architecture that   correlates with
human consciousness, and phenomenally conscious machines \cite{Gamez2008}. 
Reggia later presented a review where machine  consciousness was classified 
into   five categories based on recurring themes    on the 
fundamental issues that are most central to consciousness 
\cite{Reggia2013112}. These included a global workspace,
  information integration,   an internal self-model,  higher-level 
representation, and   attention mechanisms. With a number of challenges  
related to  definition and understanding  of consciousness, it was highlighted 
that  the way forward  to examine the inter-relationships between the five 
approaches.  Hence, it will remain very
difficult to create artifacts that truly model or support analogous
artificial conscious states.

Although various models have been discussed in detail in the reviews, we limit 
our discussion to some  of the recent models that closely relative to this 
paper. Starzyk and Prasad presented a computational model of machine 
consciousness  which was driven by competing motivations, goals, and attention 
switching through the concept of mental saccades  
\cite{starzyk2011computational}. Reggia argued that the 
efforts to create a phenomenally conscious machine have not been much less 
successful due to the  computational explanatory gap which refers to the 
inability to explain the implementation of 
high-level cognitive algorithms in 
terms of neuro-computational processing  \cite{reggia2014computational}.  It 
was highlighted  at the present time, machine 
consciousness has not presented a compelling demonstration of  phenomenal  
consciousness and further has not given  any indications for it to emerge in 
the 
future.

The social and cognitive aspects  that deal with attention and 
awareness can be helpful in further understanding certain aspects of 
consciousnesses \cite{graziano2013consciousness}. 
Graziano and  Kastner presented a hypothesis where they viewed awareness as a 
perceptual reconstruction of the attentional state. They proposed that 
 the machinery that computes information about other
people’s awareness is the same machinery that computes information about our 
own 
awareness \cite{graziano2011human}. They  proposed that 
attention and the attention schema co-evolved
over the past half-billion years and may 
have taken on additional functions such as promoting the integration of 
information across diverse domains and promoting social cognition. Their 
hypothesis further leads to a  mechanistic theory of consciousness   that 
outlined  how a brain with an attention schema may conclude that it has   
subjective awareness \cite{graziano2014mechanistic}.  In the attention schema 
theory, consciousness is viewed beyond philosophy, towards developing 
 basic properties can be engineered into machines.  It
is seen as a  fundamental part of the data processing machinery of the brain 
where  awareness is an internal model of attention. They further argued  that 
the attention schema theory provides a possible
answer to the puzzle of subjective experience whereby  the brain computes a 
simplified model of the
process and the current state of attention which is the basis of subjective 
reports \cite{graziano2015attention}. Moreover, the theory was partially based 
on the logic of model-based control motivated by how the  brain computes a 
model of the body through the body schema and uses it  in the control of the 
body. Hence, they   suggested that a simplified model of attention through  an 
attention schema would be useful in controlling attention. Lamme presented  
definitions
of visual attention and awareness that   distinguished 
between them and also explained why they are intricately related. It was 
suggested that there was overlap between mechanisms of memory and awareness
than between those of attention and awareness. Moreover, it was also 
highlighted that phenomenal experience origin from the recurrent interaction 
between groups of neurons \cite{lamme2003visual}.


 \subsection{Simulation of aspects of consciousness}
 
 Throughout modern digital history, there have been a number of developments 
in areas of artificial intelligence that mimic aspects or attributes of 
cognition and  consciousness. These developments have been made with the hope 
to replicate and automate some of the tasks that are undertaken by 
humans  given the industrial demand and  constraints  of humans on carrying out 
demanding tasks in limited time.   The replication of some of the biological 
attributes  include the feature  of learning with machine learning 
\cite{carbonell1983overview} and the attribute of reasoning and planning with 
automated reasoning \cite{wos1984automated}. The attribute  which deals with   
sensory perceptions includes the sense of hearing with speech recognition which 
covers  areas such as 
voice and speaker identification \cite{rabiner1989tutorial}. Moreover,  visual 
perception is covered through computer vision \cite{forsyth2002computer} with 
specific cases such as face \cite{turk1991face},  facial expression 
\cite{cohen2003facial}, and object recognition \cite{lowe1999object} . The 
attributes of biological motor control have been covered by  autonomous  
movement in humanoid robots  \cite{hirai1998development},  while learning to  
drive has been covered  through autonomous driving systems 
\cite{urmson2008autonomous}. Although these fields have emerged, there are a 
number of challenges that include those in computer vision and speech 
recognition, especially in dealing with noisy and dynamic environments  in 
real-world applications \cite{deng2004challenges,satyanarayanan2001pervasive}. 

The field of natural  language processing aims to make 
computer systems understand and manipulate natural languages to perform the 
desired tasks \cite{chowdhury2003natural}. It   has   been one of the    major  
 
attributes of  cognition and consciousness   \cite{halliday1999construing}.  One 
of the major breakthroughs that used 
natural language processing for cognitive computing  has been the design of   
Watson, which  is a  system capable of answering questions posed in natural 
language  developed  by   David Ferrucci 
\cite{ferrucci2010build,ferrucci2012}.  Watson won the   game of Jeopardy 
against human players \cite{markoff2011computer}. It had access to 200 million 
pages of structured and unstructured content  including the full text of 
Wikipedia. Moreover, IBM Watson was not connected to the Internet during the 
game. There are a number of applications of   Watson technology that includes 
various forms of search that have semantic properties. Specially,  
Watson has a high potential for health care  for an evidence-based clinical 
decision support system  that affords 
exploration of a broad range of hypotheses and their associated evidence 
\cite{Watson2013}. Furthermore, it can help in developing breakthrough 
research in medical and life sciences with a further focus on Big Data 
challenges . Hence, it was shown that Watson can accelerate 
the identification of novel drug 
candidates and novel drug targets by harnessing the potential of big data 
\cite{Chen2016688}.
 
 With such a breakthrough for development of Watson for cognitive 
computing, there remains deep philosophical questions from perspective of 
natural and artificial consciousness  \cite{koch2011test}. Koch   evaluated 
Watson's level of consciousnesses from perspective of integrated information  
theory of consciousnesses \cite{koch2009theory,balduzzi2008integrated} that 
views  the level of consciousness based on complexity and how integrated  the 
forms of information are in the system. Watson’s  capabilities motivated to 
further  study the philosophy, theory, and future of artificial intelligence 
based upon Leibniz’s   computational formal logic that inspired   a 'scorecard' 
approach to assessing cognitive systems \cite{Bringsjord2016}. Metacognition   
refers to a higher order thinking skills that  includes knowledge about when 
and 
how to use particular strategies for learning or for problem solving 
\cite{flavell1979metacognition}. In relation to metacognition, Watson relied on 
a skill very similar to human self-knowledge as it not only came up with 
answers 
but also generated a confidence rating for them. Therefore, Watson possessed 
elements of metacognition similar to the human counterparts in the game of 
Jeopardy \cite{fleming2014metacognition}. More recently, \textit{AlphaGo} was 
developed by \textit{Google} to play the board game \textit{Go} which  became 
the first  program to beat a professional human player without handicaps on a 
full-sized 19 $\times$ 19 board \cite{gibney2016google}. It used  deep learning  
and learned abstract information   from visual board data given by experts. 
Then it played against itself across multiple  computers  through reinforcement 
learning. Although AlphaGo has been very successful, one can argue that it 
demonstrated a very constrained aspect of human intelligence that may not 
necessarily display consciousness.

 Furthermore, ethics and morality are considered as one of the fundamental 
aspects of human consciousness. Hence, one of the future challenges will be to 
feature attributes of morality in artificial consciousness.  There have been 
questions about the moral aspects of the rise of robotic or digital  systems 
that 
will have a certain level of consciousness 
\cite{allen2000prolegomena,parthemore2013makes,Arnold2016}. Colin et. al 
proposed moral Turing test with the hope to attain  moral perfection in 
computational systems \cite{allen2000prolegomena}. Parthemore and Whitby 
questioned the requirements of a moral agent has been also 
presented a  number of conceptual pre-conditions for being a moral agent 
\cite{parthemore2013makes}.  Arnold and Scheutz argued against moral Turing 
test 
and  proposed system of verification which demands the design of 
transparent, accountable processes of reasoning  for the 
performance of autonomous systems  \cite{Arnold2016}. The issues related to 
morality would need to be integrated with systems that feature artificial 
consciousness as it could have a wide range of implications in cases when the 
system is given tasks, or in charge of making decisions that pose a danger to 
living systems. This raises further philosophical questions on the implications 
of artificial consciousness.

\section{An affective computational model  }

  \subsection{Preliminaries }

 Humans have long desired a future when advances in  robotics will 
help  solve some of the challenges facing humanity. It is well-known that 
advances in robotics and artificial 
intelligence provide potential for advances in health and  agriculture.  There 
is hope for addressing some 
of the most challenging problems  such as  the need for food, water, and 
shelter.  One 
would be glad to have a humanoid robot that can plant for the entire household 
and also help in preparation of food  and household actives. However, this will 
also give rise   to  philosophical and ethical issues.  The implementation 
of machine consciousness in technological systems  will affect human 
workforce, social behaviour and culture. 
  
 The rapid advances in   emerging technologies such as \textit{Internet of  
Things} (IoT) \cite{gubbi2013internet} is leading to increasingly large  
collection of data. IoT has the potential to improve  the  
health, transportation,  education, agriculture and other related industries. 
Apart from the
dimensionality of the data, there are other challenging   factors that 
include    complexity  and heterogeneous datasets 
\cite{doan2001recon} which makes the area of big data challenging 
\cite{Torres2016,Polsterl2016}.  
Recent success in the area of deep learning 
\cite{schmidhuber2015deep,lecun2015deep} for computer  vision and speech 
recognition tasks have given motivation for the future implementation of 
conscious 
machines. Howsoever, this raises deeper questions on the nature of 
consciousness and if deep learning with big data can lead to features that 
contribute or form some level of consciousness. Through the perspective of 
integrated information theory (IIT) 
\cite{tononi2004information,oizumi2014phenomenology}, complex structures in the 
model with feedback loops could lead to certain degrees of consciousness. 
Therefore, from the deep learning perspective, conventional convolutional 
networks 
do not fall into this category as they do not have feedback connections. 
However, if we 
consider recurrent neural networks \cite{williams1989learning}, some of the 
architectures with additional information processing would fall in the 
category of consciousness from the perspective 
of IIT.  The challenge remains in incorporating them as components form part of 
a larger model for machine consciousness \cite{sloman2003virtual}. In such 
model, deep  learning, IoT, and 
big data would replicate sensory perception. 
 
Once the simulation of input sensor organs is addressed (speech and vision), the 
challenge of an effective model of machine consciousness would be to make sense 
of the data and also provide higher level organization of knowledge obtained 
from data in which resembles thought processes and reasoning. The field of the 
semantic web has faced a similar challenge that tries to make sense of data 
from 
web content using resource description framework (RDF) which is a set of  
specifications  originally designed as a metadata data model  
\cite{berners2001semantic}. It incorporates machine learning and optimization 
through so-called web intelligence \cite{liu2003web}. They have been  
implemented in social networks and search engines \cite{breslin2007future} and 
also further enhanced by cognitive computing technologies such as  
Watson \cite{d2011watson}.

\subsection{  Affective model with simulation of natural properties    }

There has not much been done to incorporate emotional states and personality   
in models of machine consciousness. It  was not addressed as in the past due to 
the limited motivation for software  systems  and robotics which  mostly was 
aimed to address problems without taking into account of \textit{the human feel 
or touch}  which 
is recently being addressed through the field of affective computing. However, 
affective computing has yet not fully addressed its implications on machine 
consciousness. Further challenge is to  address  the hard problem   which 
refers to 
the explanatory gap of describing  conscious experience 
\cite{levine1983materialism,nagel1974like}.

We begin with the proposition where we view the human brain as 
 hardware and mind as computational 
software  \cite{pinker1999mind}. The computational software can also be viewed 
as an operating 
system that consists 
several layers and components that work coherently  as a control 
system  \cite{sloman1993mind}. We note that states in 
consciousnesses are perturbed though  emotional experiences 
\cite{ledoux1998emotional}. There has been a study on the links between  emotion 
with consciousness where 
it was suggested that emotional processing is 
important for maintaining a sense of ownership necessary for any conscious 
experience \cite{tsuchiya2007emotion}. The 
state or health of the brain has direct implications for consciousness. For 
instance, in an extreme case, someone being injured with  brain damage can 
become unconscious and enter a 
vegetative 
state \cite{bernat2006chronic}. Such natural defective states of 
consciousness resemble  damages of a computer 
hardware components such as memory and storage devices or even one of the 
processors.

\begin{figure} 
\centering
\includegraphics[width=60mm]{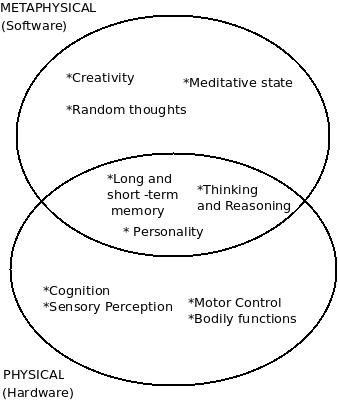}
\caption{ Overlap in states with difference in physical (hardware) and 
meta-physical 
components (software).   }
\label{fig:levels}
\end{figure}

Figure \ref{fig:levels} highlights the difference in physical 
(hardware) and metaphysical components (computational software) that form 
consciousness.  
Although the metaphysical features such as creativity and thought processes 
could  be classified as software, simulating them is difficult. For instance, 
the input for a vision-based robotic 
system would be information in terms of videos or images. The software would be 
the machine learning and data processing components that carry out tasks such 
as 
face or facial expression recognition.  Creativity, on the other hand, would 
be seen as a philosophical attribute or feature of  consciousness. Creativity  
is not just about artistic 
expressions such as fine arts or music, but about the ability to tackle 
problems from ``out of the box''. Stimulating creativity would be a very 
challenging aspect of any models of consciousness and hence we limit our 
current 
affective  model which views creativity as a black-box.

Moving on, we revisit the natural states of consciousness and  incorporate 
 components that fall between physical and metaphysical states in order to 
address the hard problems such as conscious  experience as shown 
in Figure \ref{fig:levels}. In our proposed affective model, we view the 
conscious  experience as a  state  that enables management of all the states. In 
doing so,  it  can change states depending on 
the ``present'' and ``future'' goal  as shown in Figure \ref{fig:general}. 
Furthermore, the 
affective model is developed with the following propositions. 

 \begin{itemize}
 \small
\item Proposition 1 \label{pro}:  Being conscious and unconscious are states  
of the whole phenomenon of consciousness. The model views the  sleep versus 
waking states and the major states of consciousness.

\item Proposition 2:  While being conscious, there is awareness.  
On the other hand, while being unconsciousness, there is a certain level of 
awareness and attention which are given or used in dream states. 

\item Proposition 3: 
A thinking mind is 
generation or information which can be viewed as  random walks in a network of 
information where  
there is an certain level priority to that information with  attention based 
on certain goals or emotional states. The thinking mind generates different 
types of 
thoughts 
depending on the problem at hand, the level of intelligence, depth of 
knowledge and 
experience.  
 \end{itemize}

Hence, the difference of wakeful and sleep states (conscious versus 
unconsciousness) is merely the participation of the body using motor control 
\cite{Kawato1999718}. In dreaming state, one has a virtual body which exhibits 
various actions, that are possible and also not possible (walking and flying) 
\cite{broughton1982human,nielsen1991emotions,roffwarg1966ontogenetic}. Hence, 
during dream states,   there is  conscious 
awareness. Moreover,  the person in a dream state cannot distinguish the 
difference whether the events are happening in real-life or in a dream. In 
several   levels of dream states, one may think that the situation is real which 
asks further raises questions of the difference between 
a dream and awake states. We limit our affective model 
from such philosophical interpretations, while at the same time, acknowledge 
them.  We note that there is  a hypothesis 
about the simulated universe and whether humans are subject to a grand 
simulation 
experiment   \cite{bostrom2003we}.

The  elements of pain and 
pleasure are central  driving and 
features of consciousness \cite{chandroo2004can}. In any artificial conscious 
system, their 
existence would influence in the overall emotional state of the artificial 
conscious system. The literature has the interest has been largely in trying to 
replicate 
a level of consciousness, without much interest in the future of robotics with 
 an affective or emotive features that make robots look and feel more human or 
natural. The  demand of humanoid robotics as  services to humans, the needs for 
the  human touch in 
robots will grow.  Personality is an 
attribute of consciousness that defines the way one expresses their affections 
or emotions and also handles everyday problems and situations that range in a 
wide range of settings which includes family, work and community. It is through 
one's personality, that they 
have a certain view of life that also related to moral behaviour and ethical 
constructs for behaviour. Current models of machine consciousness are 
not addressing these aspects even though they   may not address  the hard 
problems, some elements of affective behaviour could be replicated.

Personality could be seen  as an attribute of consciousness that 
grows with time 
and experience. It determines how one approaches a problem as the behaviour and 
intrinsic qualities of the person.  Although the changes in our 
emotions makes  mood that 
contributes 
to the state of consciousness,
the core identity  of consciousness remains the same, i.e we feel the same 
consciousness 
as a child or and adult although we have gone through varied learning 
experiences. This is an important aspect of the hard problem. In developing 
machine consciousness or implementing in it humanoid robotics, one can 
acknowledge the hard problem but to solve it is not necessary for attaining 
systems that have some level of affective consciousness.

Hence, such systems would have similar principle as a  parrot trying to 
replicate the conscious behaviour - which may be just repeating some words 
without  understanding it. In our analogy of humanoid robotic with affective 
consciousness, it would be carrying out a task and display behaviour that 
generate some emotion or has the human spirit or touch, but whether it is 
conscious  about it would be a philosophical discussion. Since the proposed 
model has not addressed the hard problem with any definition or discussion but 
just acknowledged its presence of conscious experience with an identity - i.e 
some state that is ``the observer'' or the ``one which experiences''. We are 
not modelling the observer as its nature has yet not fully been grasped the 
the respective scientific fields. However, in the section to follow, we will 
provide  a means for management of attributes of consciousness through an 
artificial qualia which aids the ``observer'' as the goal is to have future 
implementations of affective computational model    for robotics.

\begin{figure} 
\centering
\includegraphics[width=90mm]{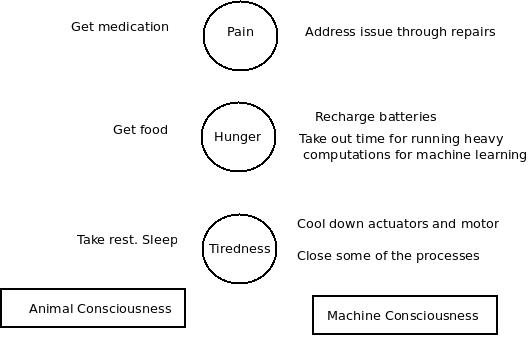}
\caption{  An illustration of animal versus machine consciousness in addressing 
some of the elements such as 'pain', 'hunger' and 'tiredness'. }
\label{fig:painmotiv}
\end{figure}

There are some intrinsic and extrinsic motivations that lead to the desire to 
reach our goals. Once a goal is established, everyday challenges  such as the 
state of pain, hunger, and tiredness remain. These states 
could be catered in the affective model of machine consciousness 
which could be helpful in addressing some of the software and hardware 
requirements. Currently, there are challenges in mobile computing, where at 
times, the battery life is running low or too many processes slow down the 
system. These could be seen analogous to challenges in animal consciousness 
such as pain 
and tiredness. Figure \ref{fig:painmotiv} provides an illustration of the 
elements that form a major part in making a close link of animal and machine 
consciousness. It shows how   the challenges could be addressed while making 
certain actions to 
achieve the goal.

We  present 
the following definitions for developing the affective computational model of 
machine consciousness. 

 \begin{itemize}
 
 \small
\item Definition 1: Any phenomenal observation is viewed  as 
information.  Computational software processes the information with knowledge 
which is either inbuilt or  gained through learning from experience or a 
combination of them. 

\item Definition 2: Consciousness  is based on attributes that have 
 qualities,   states, and   instincts.

\item Definition 3: The quality  of consciousness are those that are inbuilt, 
inherited or  born qualities such as personality, intelligence and creativity.

\item Definition 4: The states of consciousness are those that mostly change 
with phenomenal experience such as emotions, expressions and motor control.

\item Definition 5: The instinctive property of consciousness are those that 
have minimum conscious control such as body processes such as ageing, hunger 
and pain. 
 \end{itemize}
 
With the above definitions, we   address 
affective notions that include  emotional states, behaviour, and expressions 
\cite{ledoux1998emotional} while  taking into account the personality, 
knowledge and instincts as shown in Table \ref{tab:affects}. Note that the 
table forms basis for propositions based on observations only. Moreover, some 
of the identified qualities such as personality is a more rigid quality which 
may or may not change over time depending on its influence from birth. The 
property that make the  quality are merely those that we are born with or 
gained naturally, although some may change over time such as knowledge and 
creativity.

\begin{table}[h!]
\small
 \begin{tabular}{ l l l l l } 
 \hline
Property & Quality & State & Instinct & Implication\\  
 \hline\hline
 Personality & x & - & - & $D, B$, and $ M$ \\ 
  
 Intelligence & x & - & -& $D $, and $ B$ \\
  
 Creativity & x & - & -& $D $, and $B$  \\
 
 Knowledge & x & x & -&$ D, B$ and $M $\\

Memory  & x & x & - & $D, B $, and $M$\\  
 
 Extra-Sensory Percep.  & x & x& -& $ D$   \\
 \hline
 Emotions & - & x & -& $D $, and$ B$\\  
 Expression & - & x& - & $B$ \\  
 Motor Control & - & x & -& $B$ \\  
 \hline
 
Pain& -& - & x & $M$, and $B$  \\   
 Hunger & - & - & x & $M$, and $B$ \\  
 Bodily functions & - & - & x & $M$, and $B$ \\  
 \hline
 \hline
\end{tabular}
\caption{Properties of the affective computational model.  $D$ refers to 
decision 
making, $B$ refers to behaviour, and $M$ refers to  motivation.  x marks the 
presence of the attributes (Quality, State and Instinct) }
\label{tab:affects}
\end{table}

\subsection{ Problem scenarios } 

\begin{figure*} 
\centering
\includegraphics[width=150mm]{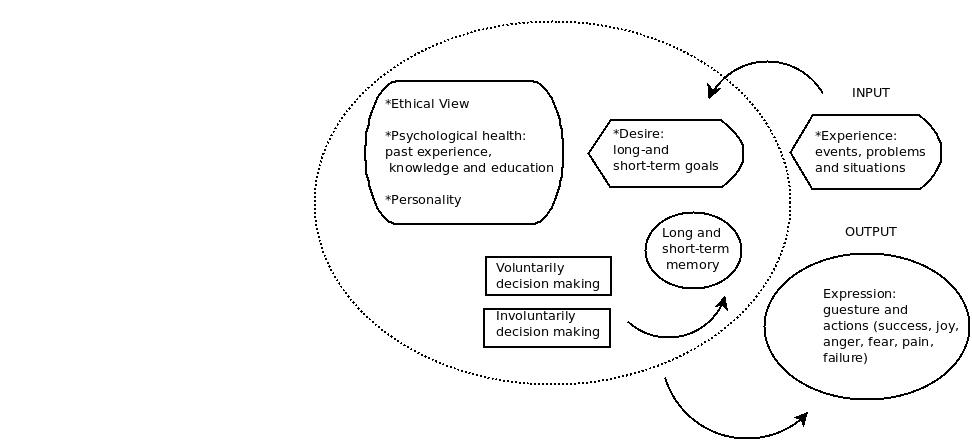}
\caption{ Output response  from input after processing through features that 
contribute to consciousness }
\label{fig:response}
\end{figure*}

Based on the prepositions in previous section \ref{pro},  we provide the 
details of the affective model and then present few problem scenarios  
that are intended to demonstrate its effectiveness.  Figure 
\ref{fig:response} shows a general view of state-based    information 
processing  based on experience which acts as input or action
while the  response acts as the  reaction given by behavior  or expression. 
Depending on the experience,  there is  an expression which would be 
involuntarily stored as either 
long or short-term memory depending on the nature of the experience. Moreover, 
there is also conceptual understanding of implications  to the observer and how 
it changes their long and  short-time 
goals. The output in terms of action or expression could also be either 
voluntary or 
involuntary. In some situations, one reacts without controlling their emotions 
while in others, one does not react in haste. A conscious decision is made 
depending on the type of 
personality, depth of  knowledge (machine learning models) from past experience 
(audio, visual and other data).

\begin{figure*} 
\centering
\includegraphics[width=160mm]{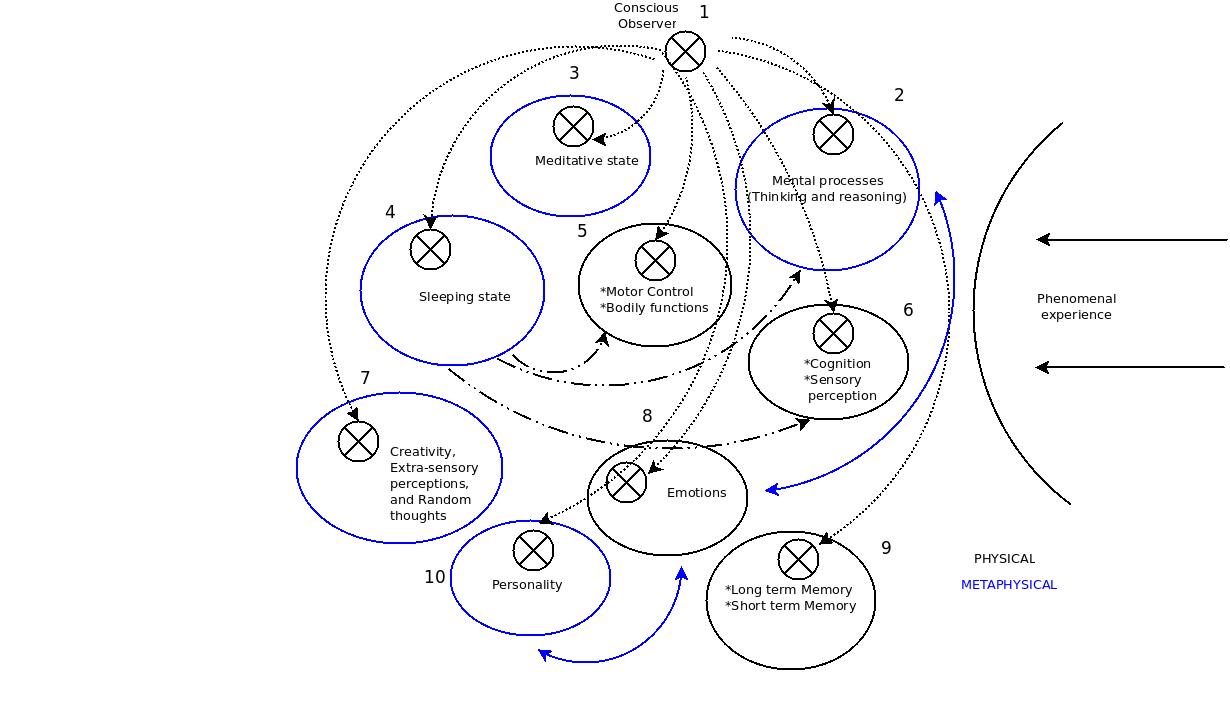}
\caption{ Note that consciousness observer is defined as the root of 
consciousness. Conscious  experience is the core, which can enter 
different states while also having the property to exist within two 
states, i.e it can self-replicate as a process, gather knowledge and update 
long 
and short-term memories, and then merge  into the root conscious observer.   The 
blue states  are  metaphysical and black states are physical.    }
\label{fig:general}
\end{figure*}

Figure \ref{fig:general} shows an over overview of the affective model of 
consciousness that is inter-related with Figure \ref{fig:response}. The states 
in Figure \ref{fig:general} shown in blue represent the 
metaphysical while those in black are the physical states. Note that by 
physical, it implies that they do have metaphysical (computational software) 
properties but 
the physical nature  influences these states.

We provide accounts of situations that require problem-solving skills which   
feature different states of consciousness. We first give the  
description of the scenario and then show how it will be tackled by the proposed 
affective model. We provide three distinct scenarios  as follows.

Scenario 1: \textit{Raman is traveling on a flight from India to Japan and has 
a 
connecting flight from Shanghai, China. His flight lands in Shanghai and he is 
required to make it to the connecting flight gate. Raman's boarding pass has 
gate information missing and since his flight landed about and hour late, he 
needs to rush to the connecting gate. Raman is not sure if he will pass through 
the immigration authority. His major goal is to reach a connecting flight gate. 
In 
doing so, he is required to gather information about his gate and whether he 
will go through the immigration processing counter. He encounters a series of 
emotions which includes fear of losing the connecting flight and hence exhibits 
a number of actions that show his emotive psycho-physical states which include 
sweating, exaggerating while speaking and even shivering due to fear. }

In order 
for Raman to successfully make it to the connecting flight on time, he will 
undergo a series of states in consciousnesses which is described in detail with 
state references from Figure \ref{fig:general} as follows.

 \begin{enumerate}
 \small

\item Exit flight and find the way to transfer desk.
  \begin{enumerate}
  \item  Search for information regarding ``transfers and arrivals'' through 
vision recognition system (State 2 and then State 6).
  \item Process information and make decision to move to the area of 
``transfers'' (State 2 and State 5).
  \end{enumerate}
\item Since information that no baggage needs to be collected was already 
given, check boarding pass for  baggage tag sticker.
  \begin{enumerate}
  \item Process visual information by checking boarding  pass (State 2 and 6)
  \end{enumerate}
\item  Confirm with the officer at transfer desk if need to go through 
migration.

  \begin{enumerate}
  \item Find and walk to transfer desk (State 2, 6, and 5)
  \item Communicate with the officer at transfer desk (State 2 and 6)
  \item Fear and emotions during communication (State 2, 5, 8, and 10)
  \end{enumerate} 
  
\item Information was given by the officer  that there is a need  to go through 
immigration booth, hence, prepare 
boarding pass and passport.

  \begin{enumerate}
  \item Rush to the immigration processing section (State 5 and 6).
  \item Wait in queue and go through a number of emotions such as fear of 
losing flight and also sweat (State 5, 6, 8, and 10).
  \end{enumerate} 
  
\item After immigration processing, find gate information and move to gate and 
board connecting flight.

  \begin{enumerate}
  \item Rush to the gate. In the process breath heavily and also sweat (State 
2, 5, and 6). 
  \item Wait at the gate with some random thoughts and then board when called 
(State 7, 8, 6, 2 and 5). 
  \end{enumerate}

\end{enumerate}

Scenario 2: \textit{Thomas is in  a mall in Singapore for his regular Saturday 
movies and shopping with friends.   Suddenly, he realizes that he can't locate 
his phone. He  brainstorms about the last few moments when he used his  phone. 
He 
goes through a series of intense emotive states that includes fear.  }

In order 
for Thomas to successfully find his phone, he will 
undergo a series of states in consciousnesses with  reference from Figure 
\ref{fig:general} as follows.

 \begin{enumerate}
  \small 

\item Thomas first informed his friends and began checking all his  pockets and 
carry bag.
  \begin{enumerate}
  \item   Check all pockets (State 5 and 6).
  \item  Inform friends and also check in carry bag (State 6, 8, 10, and 5)
  \end{enumerate}
\item  Brainstorm where was last time phone was used.
  \begin{enumerate}
  \item Ask friends when they last saw him using  the phone (State  6, 8,  and 
10).

   \item Try to remember when phone was last used (State 2 and 9).
\item Finally, take a moment of a deep breath and relax in order to remember 
(State 2, 9, and 3).

  \end{enumerate}
\item   Recalled information that phone was last used in cinema and then rush 
there to check.

  \begin{enumerate}
  \item Recalled that phone was last used in cinema (State 1, 3, 7, and 9). 
   \item Inform friends with emotive expression of hope and achievement (State 6 
and 8). 
\item Rush to the cinema and talk to the attendant with emotive state of hope 
and fear (State 5, 6 and 8).
\item Attendant locates the phone and informs (State 6).
\item Emotive state of joy and achievement (State 8).
  \end{enumerate} 
   
  
\end{enumerate}

\subsection{Artificial Qualia Manager}

We have presented affective computational model of machine 
consciousness  with 
the motivation to  replicate elements of human consciousness. This can  
exhibit  characteristics with human touch with 
emotive states through synergy with affective computing. There is a  need for 
management of components in the affective model which would help the 
property of  
consciousness experience.  Hence, there is a need for a manager for qualia. This
 could be 
seen as a  root algorithm that manages the states  with features 
that can  assign the states based on the goal and the needs (instincts) and 
qualities (such as personality and knowledge).

\begin{figure} 
\centering
\includegraphics[width=70mm]{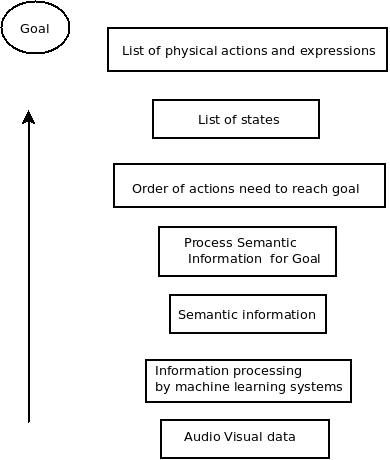}
\caption{ The journey of reaching a goal from audio visual data.  }
\label{fig:jobs}
\end{figure}

The artificial qualia manager could be modelled with 
the 
underlying principle of a security guard that
monitors a number of video feedbacks from security cameras and also has 
radio communication with other security guards and needs to follow a channel of 
communication strategies if any risks or security impeachment occurs. 
Figure \ref{fig:jobs} shows an example   that include processing through 
machine learning for semantic information which is used by the artificial 
qualia manager to assign the list of states  
needed for the goal. Similarly, the artificial qualia  manager would be 
overseeing all the status of the states and assigning jobs for reaching the 
goal through automated reasoning  in machine consciousness as given in 
Algorithm \ref{alg:qualia}.

\begin{algorithm}
\small
 \KwData{Data from sensory perception (video, audio, and sensor data) }  
 \KwResult{States for consciousness}
Initialization ( knowledge and personality)  \; 
 
 $statelist[] \leftarrow$ list of states\;
 $goal$ $\leftarrow$ gaol to reach \;
 $means[]$ $\leftarrow$  list of actions with reference to $statelist[]$ 
required to reach goal\;
  
   \While{alive}{
   
         traversestates(goal, statelist[])\;
        
        \While{goal not reached}{
              \If{challenge}{
                nominate a state\;
                attend to challenge (injury, pain, emotion) \; 
                store short-term and long-term memory\; 
              }
          
           \If{goal reached (success)}{
                output through expression (action, gesture, emotion)\;
                store short-term and long-term memory\; 
             }
          \If{goal not reached (failure)}{
                output through expression (action, gesture, emotion)\;
                store short-term and long-term memory\; 
             }
    }
        
   1. Generate random thoughts 
      based on problem and emotion \;
   2. Automated reasoning and planning  
     for states needed for future goal(s) \;
   3. Address the requirements
      to revisit failed goals \; 
     
    } 
 \caption{Artificial Qualia Manager}
 \label{alg:qualia}
\end{algorithm}

\begin{figure} 
\centering
\includegraphics[width=90mm]{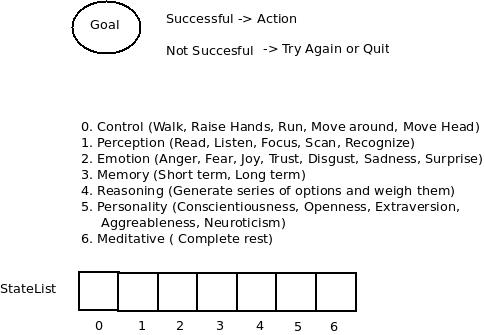}
\caption{  Affective computational  model states  for the Artificial Qualia 
Manager  }
\label{fig:states}
\end{figure}

In Algorithm \ref{alg:qualia},   the goal  and data 
from audio and visual inputs are used to determine and  effectively manage 
the sequence of  states of affective model of consciousness presented in 
Figure \ref{fig:general}. Once the goal 
is 
reached, a series  of states can  be used for expression which can include a 
set of emotions. Note that audio and visual data needs to undergo through 
processing with machine learning tools which would then output some 
information. For instance, if the goal is regarding finding date information 
for a boarding pass, then the task would be to be first to translate this 
higher level task into a sequence  of lower level tasks that would execute 
machine learning components. After these components are triggered, they would 
return information which will be used by the 
algorithm  to make further decision of states needed to reach the goal. This is 
illustrated in Figure \ref{fig:states}

 There needs to a be a property of states 
for tasks based on their importance. For instance, we give priority to 
emergency 
situations while trying to fill a goal. While fulfilling a goal, we would give 
priority to aspects such as safety and security. The goal could be similar to 
those given in Scenario 1 and Scenario 2 where Raman boards connecting flight 
and Thomas locates his phone, respectively.  

 \subsection{Implementation strategies}

The affective computational
model can feature 
multi-task learning for replicating sensory perception through recognition task 
that includes vision, sensory input for touch and smell  and auditory   tasks 
such as speech verification, speech recognition, and speaker verification. 
Shared knowledge representation would further  be used for recognition of 
objects, faces or facial expression where visual and auditory signals would be 
used in conjunction to make  a decision. Multi-task learning is  motivated by 
cognitive behavior   where the underlying 
knowledge from one task is helpful to one or 
several other tasks.   Hence, multi-task learning employs sharing of 
fundamental 
knowledge  across tasks \cite{Caruana1997,Pan2010Survey}.     

In the identification of objects, we 
learn through the experience of different senses that can be seen as a modular 
input to biological neural system 
\cite{Johnson1983}. Modular learning would help in decision making 
in cases where one of the signals is not available \cite{solomatine2006modular}. 
For instance, a humanoid 
robot is required to recognize someone in the dark when no visual signal is 
available, it would be able to make a decision based on the auditory signal. 
Ensemble learning could take advantage of several machine learning models which 
can also include deep learning for visual or auditory based recognition 
systems \cite{schmidhuber2015deep}. Ensemble learning can also be used to  
address  multi-label learning  where instances have  multiple  labels 
which is different from multi-class  learning \cite{zhang2014review}.

\begin{figure} 
\centering
\includegraphics[width=90mm]{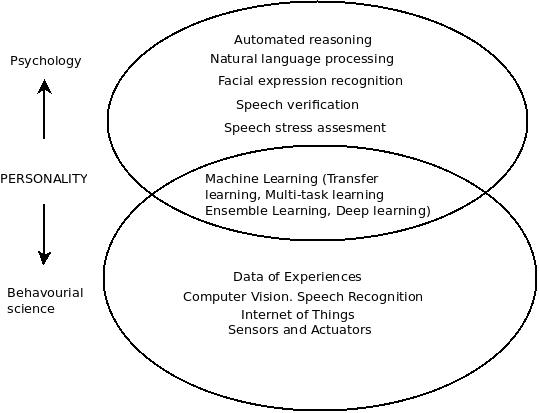}
\caption{ Use of machine learning and artificial intelligence concepts for 
implementation of machine consciousness }
\label{fig:implementation}
\end{figure}

The visual  recognition process also  relies  on information from the peripheral 
vision which is a part of the vision that occurs outside the very center of gaze 
  to make decision\cite{Yu2005Attention,Henderson2003,burgess1981efficiency}. 
Mostly, we focus our attention or 
gaze to the frontal visual system.  Similar ways of attention and focus can be 
used 
for auditory 
systems and would be helpful for advanced speech recognition systems. This is 
especially when one needs to give attention to the specific voice in a noise 
and 
dynamic environment. We naturally adjust our 
hearing to everyday situations when some parts of sensory 
inputs are either unavailable or are too noisy as trying to understand what 
someone is saying in  environments with sudden 
background noise. The feature of modularity will be very helpful in the 
development of cognitive systems for machine consciousness that need to be 
dynamic and robust. Figure \ref{fig:implementation} gives an overview of 
implementation strategies 
where machine learning methodologies are used for  
replicating sensory input through audio and visual recognition systems.

 \section{Discussion}

Although the feature of creativity, reasoning, 
self-awareness are the essential component of  consciousness, modeling them for 
aspects of machine consciousness will become the greatest challenges in the near 
future. The absence of these features will highly differentiate artificial 
systems or humanoid robots from humans and will give special qualities to the 
human workforce and hence some would argue against simulating them  
\cite{dario2001humanoids}.  
We note that self-awareness 
is a critical component of consciousness which has not been fully addressed by 
the proposed affective model which views conscious experience (observer) as 
awareness. Howsoever, these could have different philosophical interpretations 
as in the spiritual literature \cite{billington2002}, self-awareness is known to 
 emerge at higher states of consciousness or conscious  experience 
\cite{gray2004hardproblems}. The spiritual literature views   non-thinking or 
meditative state as the highest state of consciousness 
\cite{zeidan2010mindfulness}. In this state, one can  evaluate their own 
behavior and responses to problems and 
situations which can also be seen as the ability to have introspection and 
metacognition 
\cite{paivio1991dual,fleming2014metacognition,flavell1979metacognition}  . The 
challenges in machine consciousness is to incorporate 
features with fundamental models that replicate different states of 
consciousness  which align with information processing from sense organs.  
Furthermore,  intuition and creativity are also  major 
features  of consciousness and it could be argued that they form the truly 
metaphysical properties of consciousness. By metaphysical, we refer to the
aspects that transcendent thoughts or notions that cannot be defined through 
language but have an impact on emotions or a certain sense of perception 
\cite{merikle1997parallels}. It is difficult to determine whether other 
animals, who are less intelligent have 
conscious experience. Howsoever, they do have levels of cognitive problem 
solving, perception, navigation, planning, and affections. All of these 
attributes are also present in humans, and therefore, any artificial conscious 
system that exhibits  these properties will face the same challenges or 
philosophical questions if animals 
have consciousness or conscious experience.

It is important to realise the 
potential of animal consciousness as it can motivate models for 
consciousness that full the gaps in models for human consciousness. In 
simulation or the need to implant certain level of consciousnesses to robotic 
systems, it would be reasonable to begin with  animal level where  
certain tasks can be achieved. For instance, a robotic system that can 
replicate cognitive abilities and level of consciousness for rats can be used 
for some tasks such as burrowing holes, navigation in unconstrained areas for 
feedback of videos or information, in disasters such as earthquakes and 
exploration of remote places, and evacuation sites.

Deep learning, data science and analytics can further help in contribution 
towards 
certain or very limited areas of machine consciousness. This is primary to 
artificially replicate areas of sensory input such as artificial speech 
recognition and artificial vision or perception. Howsoever, with such 
advancements in artificially replicating sensory perceptions, one encounters 
further challenges in developing software systems that oversee or synchronise 
different aspects of perceptions that lead to a consciousness state.  
Howsoever, 
to reach a state of natural 
consciousness will be difficult for machines as creativity and self-awareness 
is 
not just biological, but also considered spiritual which is challenging to 
define.

With the rise of technologies such as IoT, sensors could be used to 
replicate biological   attributes such as pain, emotions, feeling of strength 
and tiredness. However, modelling
these attributes and attaining same behaviour in humans may 
not necessarily mean that the   affective model of consciousness would solve 
hard problem that  enables conscious 
experience. However, at least the model  would be seen to exhibit conscious 
experience that will be similar to humans and other animals. Such an affective 
model, with future implementations could give rise to household robotic pets 
that would have or could develop emotional relationship with humans. We must be 
careful about affective model when in giving autonomous control or decision 
making through simulated emotional behaviour. Humans are well known to be poor 
decision makers when in emotional states which also resort to level of 
aggression and violence. Therefore, simulation of affective states need to take 
into account of safety and security for any future robotic implementations that 
assist humans.

The proposed affective model has not considered any difference between 
conscious experience during sleep and waking state from the perspective of 
awareness \cite{lamme2003visual}. This is due to the difference in the 
definition of awareness  from the sleep and waking state 
\cite{tassi2001defining}. We note that artificial systems do not need elements 
such as the sleep state as its a property of a biological nervous system where 
sleep is required. Moreover, during the sleep state, dreams are persistent and 
their 
importance has been an important study in psychology 
\cite{nielsen1991emotions}, 
but may not have  implications for    the affective model of consciousness. 

 

\section{Conclusions and Future Work}

  The paper presented an affective computational  model for machine 
consciousness with the 
motivation  to feature the emotive attributes which give a more human-like 
experience for artificial systems. The affective model can 
become the foundation for developing  artificial systems  that can assist humans 
while appearing as natural as possible.  

The challenges lie in further refining specific features such as personality and 
creativity which are psycho-physically challenging to 
study and hence pose limitations to the affective model of consciousness. 
Howsoever, the proposed effective model can be a baseline and motivate the 
coming decade of simulation and implementation of machine consciousness for 
artificial  systems such as humanoid robots.   The simulation for 
affective model of consciousness with the features of artificial qualia 
manager can 
also be implemented with the use of robotics hardware. In their absence, 
simulation can also be implemented through collection of audiovisual data and 
definition of certain goals. The affective model is general and does not only 
apply to humanoid robots, but can be implemented in service application areas 
of software systems and technology.

Future directions can be in areas of  artificial personality and  
artificially creative 
systems. These can be done  by incorporating advancing technologies such as 
IoT, semantic web, cognitive computing and machine learning, along with  
artificial general intelligence.

 \section{References}

  \bibliographystyle{IEEEtran}
 
 
\bibliography{rr,mergedApril2015,feature,cons}

\end{document}